\documentclass[12pt]{a_t}
\usepackage{graphicx}
\usepackage{cmap}
\usepackage{url}

\begin{document}  

\year{2024}
\title{Smart Routes: a system for development and comparison of algorithms for solving vehicle routing problems with realistic constraints}%

\authors{A. Soroka (andrew.soroka@student.msu.ru),\\
G. Mikhelson (mikhelson.g@gmail.com)\\
(Moscow State University, Faculty of Computational Mathematics and Cybernetics, Moscow),\\
A.~Mescheryakov (mesch@cosmos.ru)\\
(Moscow State University, Faculty of Computational Mathematics and Cybernetics, Moscow, Space Research Institute of RAS, Moscow),\\
S.~Gerasimov (gerasimov@cs.msu.ru)\\
(Moscow State University, Faculty of Computational Mathematics and Cybernetics, Moscow)}

\maketitle

\begin{abstract}
The problem of route optimization with realistic constraints is becoming extremely relevant in the face of global urban population growth. While we are aware of approaches that theoretically provide an exact optimal solution, their application becomes challenging as the problem size increases because of exponential complexity. We investigate the Capacitated Vehicle Routing Problem with Time Windows (CVRPTW) and compare solutions obtaining by exact solver SCIP \cite{BestuzhevaEtal2021OO} with heuristic algorithms such as LKH, 2-OPT, 3-OPT \cite{lkh2000}, the OR-Tools framework \cite{perron2011operations}, and the deep learning model JAMPR \cite{falkner2020learning}. We demonstrate that for problem of size 50 deep learning and classical heuristic solutions became close to SCIP exact solution but requires less time. Additionally for problems with size 100, SCIP exact methods $\sim$ 13 times slower that neural and classical heuristics with the same route cost and on $\sim$ 50\% worse for the first feasible solution on the same time. To conduct experiments, we developed the Smart Routes platform for solving route optimization problems, which includes exact, heuristic, and deep learning models, and facilitates convenient integration of custom algorithms and datasets.
\end{abstract}

\section{Introduction}
    The Vehicle Routing Problem (\textbf{\textit{VRP}}) is a class of transport logistics problems that aims to reduce the cost of transport resources, route expenses, and cargo delivery time for a group of customers. In real-time scenarios, optimizing routes while considering various constraints is an issue for most companies. As the number of cities and customers increases, it becomes necessary to develop a solution that optimally utilizes allocated resources while maintaining service quality. A shorter route enables faster delivery for customers and allows more time to deliver goods to other customers.

    In logistic problems, the \textit{VRP} is often considered with various constraints that need to be taken into account when constructing the routes. The most popular ones are constraints on \textit{customer time windows and service times} (TW) and \textit{vehicle capacities} (C). Currently, there are significant challenges in developing and modifying algorithms capable of solving problems with increasing dimensions while striking a balance between solution quality and search time based on the number of imposed constraints. Additionally, there is no unified platform that allows users not only to solve problems using built-in algorithms and experiment with them but also to add their own algorithms, data reading and processing methods, etc., without changing the system's architecture.

    The contribution of this paper is the following:

    \begin{itemlist}
        \item A platform for route problems optimization;
        \item A comparison of the effectiveness of common route optimization methods.
    \end{itemlist}

    Our main focus is on the development of an innovative system called Smart Routes and an in-depth comparative analysis of different methodologies in the context of problematic constraints. We have studied heuristics, deep reinforcement network and exact method (SCIP), taking into account the limitations discussed in problems of different dimensions. Our study concludes with valuable information about the efficiency and quality of these solutions, which are primarily assessed using two main metrics: the solution optimization time and the final route cost determined by the objective function.

    This paper reveals the architecture of our Smart Routes system designed to optimize route problems. The system seamlessly integrates exact and heuristic approaches, as well as deep reinforcement network methodologies, designed to solve the vehicle routing problem (VRP) in its various forms and modifications. So, it is useful for both experienced transport logistics experts and those less experienced in the field, providing a one-stop platform for testing new ideas. Consequently, this study not only delves into the ongoing search for the most efficient algorithmic approach to solving the Vehicle Routing Problem with Time Windows (CVRPTW) --- a subcategory of VRP with popular and practical limitations—but also introduces an innovative tool ready to solve this problem comprehensively and efficient.

    The article is structured as follows: Section 2 provides a review of recent literature containing both classical and deep reinforcement approach to route optimization. Sections 3 and 4 discuss the models considered and the developed Smart Routes system, respectively. Section 5 describes the data on which the experiments were conducted. In Section 6, we present the obtained results, and Section 7 contains the conclusions drawn from the conducted study.

    $$
    failure\_rate = \frac{number\_of\_unsolved\_tasks}{total\_number\_of\_tasks} \times 100 \%
    $$

\section{Related work}
    This section considers three groups of algorithms capable of solving the problem. We note that although the problem has a rich history, there are many studies comparing different approaches to solving the route optimization problem. Most papers tend to suffer from a lack of exact  approaches in comparison and either limited size of the problems considered, sometimes lack of constraints. We try to cover all the requirements in order to leave a complete picture of the applicability of classical approaches.

    \subsection{Heuristic algorithms}
        The first group of algorithms is known as \textbf{\textit{heuristic algorithms}}, which are commonly used for such problems. Within this group, two distinct classes can be identified: \textit{constructive heuristics} and \textit{metaheuristics}.

        \textbf{Constructive heuristics} are algorithms that iteratively build feasible solutions by adding components one at a time, such as the \textit{nearest neighbor heuristic}, \textit{insertion heuristic}, and \textit{sweep heuristic}. While they do not guarantee optimal solutions, they can quickly generate high-quality solutions.

        \textbf{Metaheuristic algorithms} are more general optimization methods that explore the solution space using randomization and heuristics, such as \textit{simulated annealing}, \textit{genetic algorithms}, and \textit{ant colony optimization}. They may find better solutions than constructive heuristics but often require more computational resources.

        The choice of these algorithms depends on the specific task and its application. Constructive heuristics are commonly used in practice due to their speed and efficiency in quickly obtaining high-quality solutions. On the other hand, metaheuristic approaches are more flexible and can be applied to a wider range of optimization problems, however with higher computational costs. Constructive heuristics are faster to implement and can provide effective solutions for certain problems.

    \subsection{Exact algorithms}
        The next group consists of \textbf{exact algorithms}, including linear programming algorithms. These algorithms deal with problems in which the objective and constraint functions are linear. Linear programming problems have been extensively studied, and their solution properties are well-known.

        \textit{Linear programming problem} (\textit{\textbf{LP}}) is an optimization problem expressed in standard form as follows:
        \begin{equation}
            \max {c^{\mathrm{T}} x \mid Ax \leq b, x \geq 0}
        \end{equation}

        Here, $A\in R^{m,n}$ represents the technological matrix, $b\in R^m$ is the resource vector, $c\in R^n$ is the price vector, and $x\in R^n$ is the unknown vector. If several or all variables in the vector \textit{\textbf{x}} are integers, the problem is referred to as \textit{MILP} (\textit{\textbf{Mixed Integer Linear Programming}}).
        
        There are several algorithms that can solve \textit{MILP} problems:

        \begin{enumerate}
            \item \textit{Branch\&Bound} \cite{laporte1983} is a widely used algorithm for solving MILP problems. It divides the problem into subtasks called nodes and solves each node using linear programming. The algorithm branches the node into subnodes by adding constraints until an integer solution is found or infeasibility is proven.

            \item \textit{Cutting Plane} \cite{cook1999} iteratively adds valid linear inequalities, called cuts, to exclude non-integer solutions in a problem formulation. Cuts are generated by solving the \textit{LP}-relaxation of the problem. The algorithm continues until an integer solution is found or infeasibility is proven.

            \item \textit{Branch\&Cut} \cite{augerat1995} is a combination of the \textit{Branch\&Bound} and \textit{Cutting Plane} algorithms. It generates cuts to eliminate non-integer solutions and branches nodes to create subnodes. This algorithm is typically more efficient than using Branch\&Bound or Cutting Plane separately.
        \end{enumerate}

        In general, the choice of algorithm depends on the specific problem and the requirements for solution quality and search time. Some \textit{MILP} solvers, such as \textit{CPLEX} \cite{cplex2009v12}, \textit{SCIP} \cite{BestuzhevaEtal2021OO}, and \textit{Gurobi} \cite{gurobi}, implement several of these algorithms and automatically select the most appropriate one for a given problem instance.

        The main challenge with linear programming problems lies in their high dimensionality, often involving thousands of variables and constraints. The memory usage and solution time can increase exponentially as integer variables are added. Therefore, heuristics have been developed to find approximate optimal solutions in less time. For complex problems, heuristic approaches often provide the best trade-off between solution quality and computational time.

    \subsection{Deep learning and reinforcement learning algorithms}

    The first deep learning model proposed for solving VRP was introduced by Nazari et al. \cite{nazari2018reinforcement}, who adapted the Pointer Network (PtrNet) from Vinyals et al. \cite{vinyals2015pointer} to work with CVRP. Nazari et al. \cite{nazari2018reinforcement} completely abandoned the original RNN encoder part of the model and replaced it with a linear layer with shared parameters. A more recent algorithm, AM (attention model) by Kool et al. \cite{kool2018attention}, replaced this architecture with an adapted transformer model using attention \cite{vaswani2017attention}. A direct improvement to this model is the JAMPR approach proposed by Falkner et al. \cite{falkner2020learning}, where the authors added additional fully connected networks for the current path and truck positions. This addition allowed the algorithm to successfully solve the \textit{CVRPTW} problem. Chen and Tian \cite{chen2019learning} propose an RL-based approach that iteratively selects a region on the graph and then selects and applies established local heuristics. This approach was further enhanced by the destruction operator introduced by Lu et al. \cite{lu2019learning}. The latest attempt to use deep learning for partitioning a set of points into subproblems and solving them using a black-box solver was proposed by Li et al. \cite{li2021learning}. The authors presented two approaches: regression-based prediction of potential improvement in the final cost and classification for the best subproblem. By reducing the dimensionality and utilizing classical metaheuristic approaches in each subproblem, the authors were able to achieve good results on high-dimensional problems (over 1000 points). We have presented and explored in detail the applicability of our deep reinforcement approach to solve the route optimization problem with realistic constraints, showing how a fast suboptimal solution can be obtained using learned neural heuristics in the latest article \cite{soroka2023deep}.

\section{Algorithms and models}

    \subsection{Classic approaches}
    
        While considering classical heuristic approaches, the decision was made to focus on a group of algorithms known as \textit{local search}. These algorithms provide a good balance between solution quality   and search efficiency in real-world problems and also serve as fundamental components for metaheuristic and genetic algorithms \cite{groer2010library}.
    
        One of the best classical heuristic approaches that was chosen as the main heuristic algorithm is the \textit{Lin-Kernighan heuristic} \cite{lkh2000}. This algorithm belongs to the class of local optimization algorithms. It operates by performing exchanges or moves, referred to as \textit{opt}, that transform one route into another. Starting from an initial feasible route, the algorithm recurrently performs exchanges that reduce the length of the current route until a route is reached where no further exchange can lead to improvement. This process can be repeated multiple times from initially generated routes in a randomized manner. The algorithm assumes the existence of an initial route partitioning, and then iteratively improves upon the existing approximation. The improvement method involves exchanging vertices within each sub-route individually.
    
        SCIP \cite{BestuzhevaEtal2021OO}, an open-source software, is a powerful optimization tool specifically designed for solving mixed-integer programming problems \cite{mip2013, mip2020}. It is well-suited for addressing complex optimization challenges encountered in logistics, planning, and production scheduling. SCIP employs a range of methods and algorithms, including Branch\&Bound, Cutting Plane methods, Constraint Propagation, Heuristics, Decomposition methods, and Integer Programming. These techniques involve dividing the problem, adding constraints, utilizing inference, applying rules of thumb, and solving integer programming problems to find optimal solutions. In summary, SCIP is a comprehensive optimization software that utilizes diverse algorithms and methods to effectively and accurately solve complex optimization problems.
       
        In addition to SCIP, we selected OR-Tools (Operations Research Tools) \cite{perron2011operations}, a highly regarded framework renowned for its effectiveness in solving VRP (Vehicle Routing Problem) challenges. OR-Tools is a versatile software designed to tackle combinatorial optimization problems, equation and inequality solving, and scheduling and routing problems. It offers a diverse array of optimization methods and algorithms, including \textit{Linear Programming (LP) methods}, \textit{Integer Programming (IP) methods}, \textit{Discrete Optimization methods}, and \textit{Equation and Inequality solving methods}. Notably, OR-Tools excels in Routing and Scheduling, providing algorithms for solving various routing problems such as the traveling salesman problem and vehicle routing problem. These methods encompass both heuristic approaches and exact techniques like local search and metaheuristic algorithms. With its flexibility and comprehensive range of methods, OR-Tools proves to be a powerful tool for addressing optimization problems across domains such as logistics, planning, and transportation.

    \subsection{Deep reinforcement learning models}
        We modified the JAMPR model \cite{falkner2020learning} and use as our deep reinforcement approach \cite{soroka2023deep}. JAMPR model is a modification of the attention model (AM) \cite{kool2018attention} by Kool et al., which utilizes an attention-based encoder-decoder architecture \cite{vaswani2017attention}. Both models consider the route optimization problem as a sequential decision-making problem, modeled as a Markov decision process and solved using reinforcement learning. The problem solution is built incrementally by creating routes one node at a time. The current solution, route, and unvisited nodes are interpreted as the state, and the index of all unvisited nodes available for addition to the current route is treated as the actions. 

\begin{figure}
        \centering
        \includegraphics[width=0.45\textwidth]{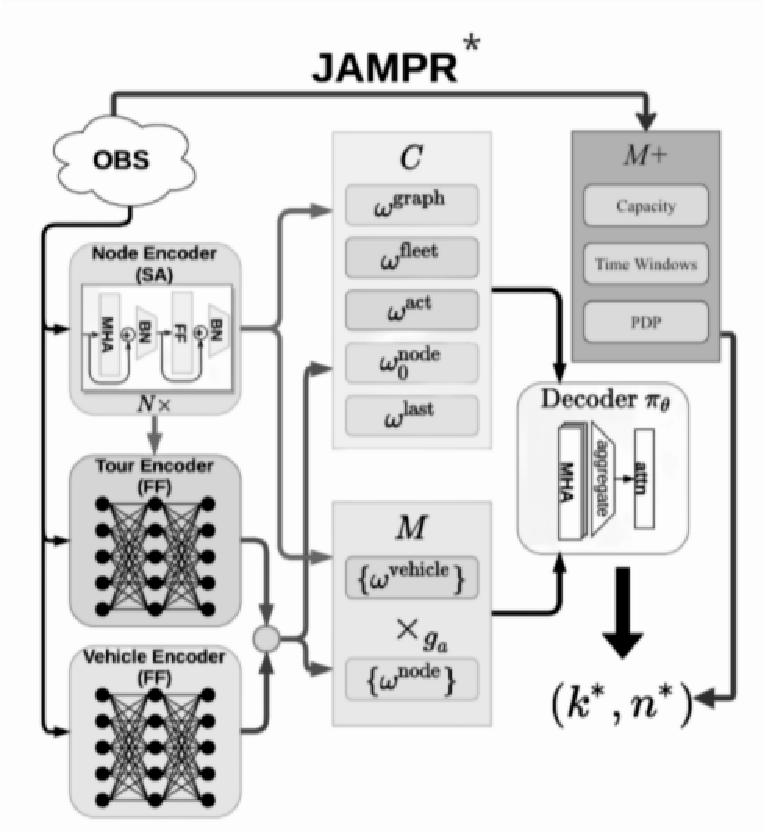}
        \caption{The JAMPR deep RL architecture modified to solve CPDPTW problems. Added block $M+$ of learnable masks applied to output of decoder (policy $\pi_\theta$). }
        \label{fig:jampr}
        \end{figure}

        We added to the JAMPR approach trainable matrices (M+ block from \ref{fig:jampr}, separate for each constraint. We apply them to the result of the decoder, thus modifying the policy($\pi_\theta$). The final result of the network operation is a modified policy of size (k*, n*). The optimization process consists of minimizing the cost of the route, taking into account missed clients (soft setting):

        \begin{equation}
        Q = \sum_{j=1}^{m} \sum_{i=1}^{n} c_{ij} \cdot x_{ij} + \lambda \sum_{i=1}^{n} z_i 
        \label{eq1}
        \end{equation}

        where clients $i$, $i=1,\ldots,N$, vehicles $j$, $j=1,\ldots,M$, $c_{ij}$ - cost of route for $j$ vehicle to $i$ client and $z_i$ - binary parameter if we missed client.
        
        Workflow of model: the encoder receives the features $x_i$ of each node $i$ (coordinates, cargo weights, time windows, etc.) and encodes them into a hidden linear vector $\tilde{x_i} \in R^{d_{emb}}$ of size $d_{emb}$. Then, the decoder model computes the attention query for each $\tilde{x_i}$ with respect to a specific context $C^{(t)}$ at decoding step $t$, in order to obtain scores for all nodes that can be added to the current route. Here, the context incorporates implicit graph embedding of the problem and additional problem-specific information such as the depot node index, the last node added to the current route, and the remaining capacity. The obtained scores are then either used in a greedy selection procedure, where the node with the highest score is always chosen, or transformed using softmax into a distribution used for sampling. In general, the encoder-decoder model represents a policy $\pi(i^{(t+1)} \vert C^{(t)},x;\theta)$ with trainable parameters $\theta$. You can see full $JAMPR^*$ architecture from original paper \cite{falkner2020learning} on figure \ref{fig:jampr}.

\section{System \textit{Smart Routes}}
    \subsection{Requirements}
        System should meet the following functional \textbf{basic requirements}:
        \begin{itemlist}
            \item \textbf{Solving VRP with popular constraints.} This involves solving VRP with common and important constraints such as vehicle capacity and time windows for delivery. As the system expands, new constraints will be added.
            \item \textbf{Solving large-scale problems ($\sim$1000 points).} Solving large-scale VRP presents challenges as the number of customers to be visited increases, making it difficult for algorithms to find solutions and leading to longer search times.
            \item \textbf{Using real-world data.} While our work compares algorithms using artificially generated data, the proposed system also includes a feature for processing real-world data in a specific format.
            \item \textbf{Conducting experiments with classical heuristic, exact, and deep reinforcement network approaches.} Problem of identification of the algorithm type that is best suited for solving VRP remains relevant Therefore, our system provides the opportunity to experiment with algorithms from different classes.
            \item \textbf{Interactive usage of the system.} To facilitate the use of built-in algorithms, an intuitive interface is provided for adjusting parameters and interacting with the system.
            \item \textbf{Visual interpretation of solutions.} Implementing algorithms can sometimes make it difficult to determine the correctness and optimality of solutions. Visualizing the solution simplifies this verification process by providing a representation of the displayed route.
            \item \textbf{Comparison of quality indicators.} When comparing multiple algorithms, it is useful to analyze the differences in the quality of the obtained solutions. Graphs demonstrating the decrease or increase in route cost over a certain period of time are particularly useful and convenient for conducting these experiments.
        \end{itemlist}
        
    \subsection{Architecture}
        The system architecture is shown on Figure \ref{sec:DescriptionOfThePracticalPart:fig:ExperimentalStand}. Let us describe the system's main flow:
        \begin{figure}[!h]
            \centering
            \includegraphics[width = 1.0\linewidth]{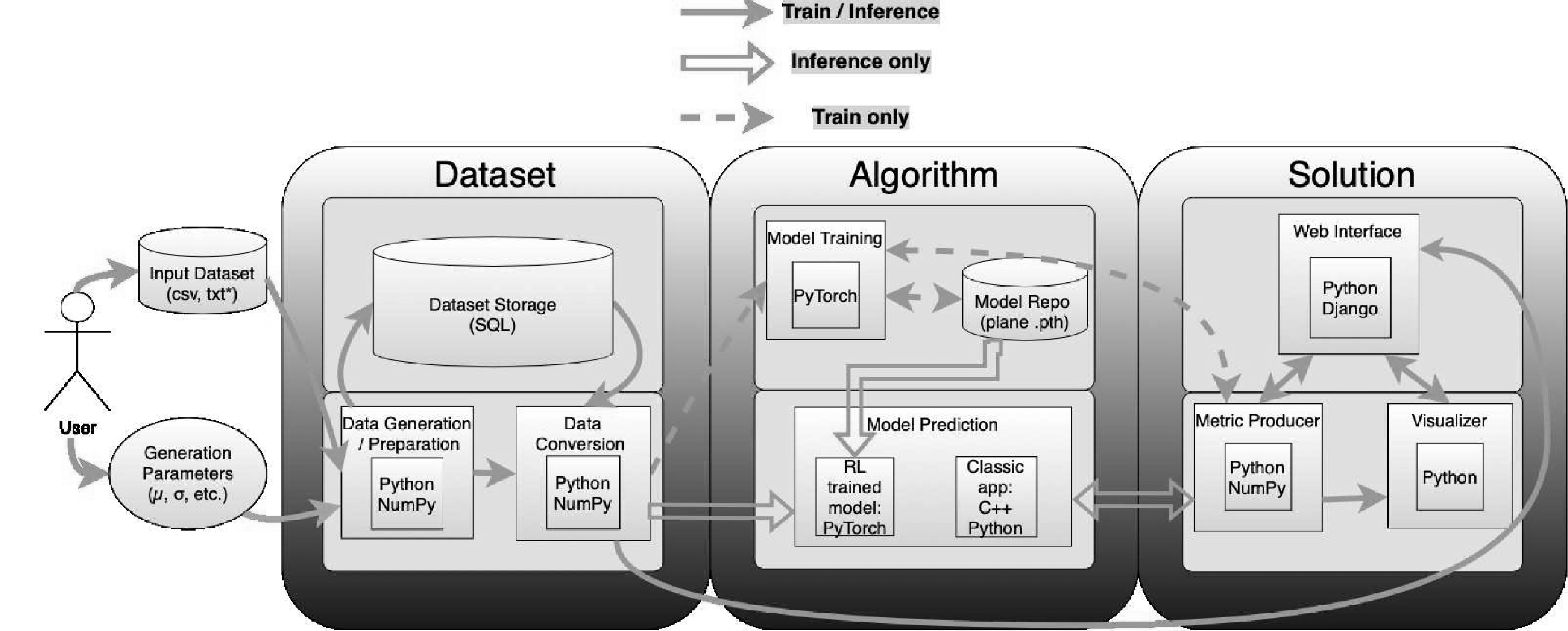}
            \caption{Smart Routes System architecture. Three main blocks: 1) Dataset module is an entry point for all data manipulations, also stores prepared datasets and allows loading third-party data in a specified format; 2) Algorithm module responsible for train and evaluate all available algorithms on prepaired datasets. There is possibility to add user algorithm via special interfaces; 3) Solution module produces metrics, visualizes them and offers web interface for communication with system. }
            \label{sec:DescriptionOfThePracticalPart:fig:ExperimentalStand}
        \end{figure}

        The user installs the system, strictly following the documentation. Upon launching it, the user is redirected to a web browser where a web page is displayed, prompting them to enter the necessary parameters.
        \begin{enumerate}
            \item \textbf{Data and Parameters.}
            \begin{enumerate}
                \item The user selects the problem they intend to solve: \textit{CVRP} (Capacitated Vehicle Routing Problem), \textit{VRPTW} (Vehicle Routing Problem with Time Windows), \textit{CVRPTW} (Capacitated Vehicle Routing Problem with Time Windows);
                \item Name of the solving algorithm is chosen;
                \item A time limit is set for solving one task;
                \item The option to solve tasks sequentially or in parallel on the available number of processors is chosen;
                \item A dataset is loaded in a specific format.
                \item Parameters for generating datasets are specified if needed: 1) $\mu$ --- mean value; 2) $\delta$ --- variance; 3)  $capacity_{min}$ - minimum consumer demand value; 4) $capacity_{max}$ - maximum consumer demand value; 5) $n_{samples}$ --- number of tasks; 6) $n_{customer}$ --- number of customers in each task; 7) $service_{window}$ --- right time window for the depot; 8) $service_{duration}$ --- customer service time.
            \end{enumerate}
            \item \textbf{Dataset.}
            The user-defined parameters and loaded data are passed to the \textit{Data Generation/Preparation} block, where the data is either artificially generated based on the user's specified parameters or stored in variables. The user datasets are then saved in a database or immediately passed to the \textit{Data Conversion} block, where the data is transformed by converting it to the required types.
            \item \textbf{Algorithm.} 
            This block receives the preprocessed data and user-specified parameters. The system determines which class of problems the selected algorithm belongs to: \textit{classic\_heuristics}, \textit{exact\_methods}, \textit{neural\_methods}.

            As the result, for each submitted task, the user obtains the final route, i.e., a list of customers and its cost.
            \item \textbf{Solution.} 
            The results obtained from the previous block are passed to the \textit{Metric Producer} block, which generates an overall graph with metrics for the user-selected algorithms, demonstrating the decrease in route cost within the specified time limit. The results are also passed to the \textit{Visualizer} block, which visualizes the complete graphical route based on the obtained list of customers.
        \end{enumerate}
        Upon completion, the system displays two generated graphs, the final route cost, and the list of customers in the order of their visitation on the initial web page.

    \subsection{Advantages}
    Our platform has a key feature that allows users to add their own algorithms and models, giving it a significant advantage over other frameworks because:
    \begin{enumerate}
        \item Understanding the framework's toolkit can take a lot of time. For example, implementing the CVRPTW model using SCIP software required a considerable amount of time to understand how to correctly implement constraints, define the model architecture, and preprocess input data.
        \item Software packages such as OR-Tools, SCIP, Gurobi, CPLEX only allow building models that act as exact methods. As the dimensions of the problem instances increase, the solution search time also grows, which can become critical.
    \end{enumerate}
    
    In our system, users can integrate their algorithms by inheriting from base classes and implementing the necessary methods. The built-in algorithms can be applied to all types of approaches capable of solving VRP.
    
    In addition our system also includes two convenient features. The first feature is result visualization, which is essential for evaluating the performance and suitability of the developed approach for the problem at hand. It automatically visualizes the final route within a specified time frame, allowing users to assess the solution. The second important feature is the API, which allows users to easily utilize the built-in approaches without integrating their own algorithms. By launching the server, users can upload a dataset, select an approach, and input additional parameters through a browser page. They will then receive the overall route cost, the route itself, and its interpretation.
    
    Thus, these platform features simplify the development, testing, and experimentation of VRP solution approaches and demonstrate a significant advantage over existing software.

    \subsection{Comparison of Smart Routes and Classical Frameworks}
    \label{compareframeworks}
    Currently, there are some frameworks that provide tools for solving the Vehicle Routing Problem (VRP) with various constraints. The most popular ones include OR-Tools, Gurobi, and SCIP.
    
    The table below shows the types of algorithms supported by these frameworks, denoted by + and - :
    
    \begin{table}[!h]
        \centering
        \begin{tabular}{|c|cccc|}
            \hline
            Algorithm Type              & \multicolumn{4}{c|}{Framework}                                  \cr \hline
            \multicolumn{1}{|l|}{} & \multicolumn{1}{c|}{SCIP} & \multicolumn{1}{c|}{Gurobi} & \multicolumn{1}{c|}{OR-Tools} &
            \multicolumn{1}{c|}{Smart Routes} \cr
            \hline
            Exact Approaches & 
            \multicolumn{1}{c|}{+} & \multicolumn{1}{c|}{+} & \multicolumn{1}{c|}{+} & \multicolumn{1}{c|}{+} \cr 
            \hline
            Heuristic Approaches & \multicolumn{1}{c|}{+} & \multicolumn{1}{c|}{+} & \multicolumn{1}{c|}{+} &
            \multicolumn{1}{c|}{+} \cr
            \hline
            Deep Reinforcement Approaches & \multicolumn{1}{c|}{---} & \multicolumn{1}{c|}{---} & \multicolumn{1}{c|}{---} & \multicolumn{1}{c|}{+} \cr
            \hline
            Visualization of Results & \multicolumn{1}{c|}{---} & \multicolumn{1}{c|}{---} & \multicolumn{1}{c|}{---} & \multicolumn{1}{c|}{+} \cr 
            \hline
            Adding Custom Algorithms & \multicolumn{1}{c|}{---} & \multicolumn{1}{c|}{---} & \multicolumn{1}{c|}{---} & \multicolumn{1}{c|}{+} \cr
            \hline
        \end{tabular}
    \end{table}
    It is important to note that OR-Tools and SCIP may not provide the same level of performance and scalability as Gurobi when solving very large-scale optimization problems.
    
    Our proposed system differs from the aforementioned software in several aspects. In addition to covering all types of algorithms for solving the VRP, Smart Routes allows for the integration of custom models, showcasing its flexibility without compromising the overall architecture. We have implemented visualization features to create performance charts and visualize the final routes. Furthermore, during the system installation, OR-Tools and SCIP are automatically installed, making it easier to conduct experiments comparing user models with models implemented using the toolsets provided by these frameworks. There is no need to develop models or algorithms for solving custom logistics problems, as our system already offers built-in implementations. Moreover, our system provides a user-friendly interface that simplifies the workflow and makes it a convenient tool for users.

\section{Data}
    The algorithms considered in this research aimed to solve the \textit{CVRPTW} problem. The interest is to observe the behavior of these approaches when multiple constraints are present simultaneously. In particular, the problem was solved in the SOFT settings, which allow skipping points that cannot be visited. In such cases, these points are excluded from the final route and incur a penalty, represented by an increase in the \textit{missed\_nodes} variable in the cost formula.

    For the \textit{CVRPTW} problem, suitable instances were selected from the well-known reference set of \textit{Solomon} based on the R201 statistics \cite{solomon1987algorithms}. The truck volumes were given as $Q_{50} = 750$ and $Q_{100} = 1000$ for problem sizes of 50 and 100, respectively. The total time horizon is defined as $[a_0, b_0]$, where $a_0 = 0$ for all examples, and the right boundary $b_0$ varies depending on the problem size, specifically 1000 for 50 and 100 points. Additionally, the service duration $h_i$ for each point is uniformly set to 10.
    
\section{Experiments}
    Artificial datasets were generated for the experiments, consisting of 100 examples with dimensions of 50 and 100 points. Each problem was assigned a specific time limit for solving using the algorithms: 100 seconds for 50-point problems and 200 seconds for 100-point problems. Since exact methods involve a complete search of possible solutions, they required 10 times more time for each problem to obtain any solution, even if it is not optimal. Therefore, the exact method was given 1000 and 2000 seconds for 50-point and 100-point problems, respectively.

    The experiments were conducted using the developed system Smart Routes, which simplified the process of setting parameters when running different algorithms and tasks. To save time during the experiments, the tasks were parallelized on all available processors of the machine.

    All experiments were performed on a server with NVIDIA Tesla A40 GPU and Intel(R) Xeon(R) Gold 6226R CPU @ 2.90GHz with 16 virtual cores.

    \subsection{Main Results}
\begin{figure}[!h]
        \centering
        \includegraphics[width=0.83\linewidth]{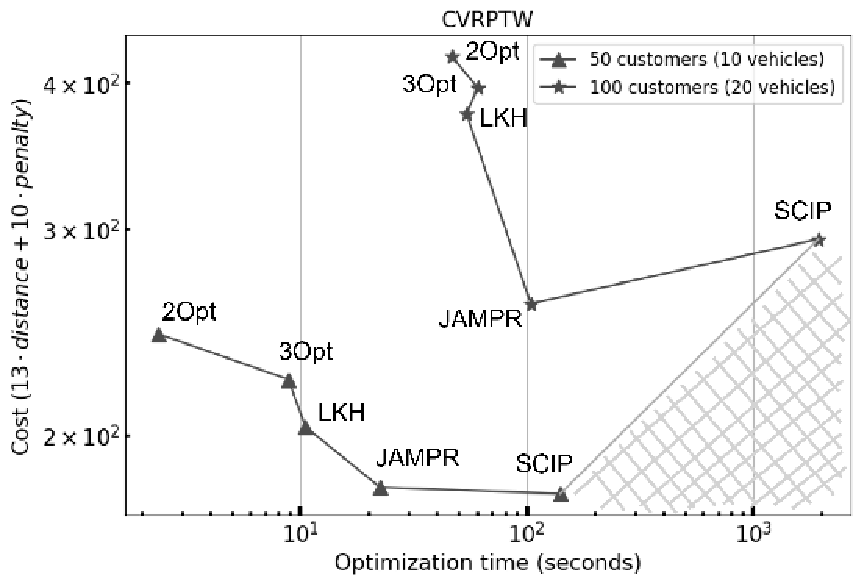}
        \caption{Summary result of classical approaches, where each dot denotes the result of a specific algorithm on a given problem size. The horizontal line indicates the time required for the algorithm to achieve the best cost of the path, the vertical line indicates the best cost achievable by the algorithm. The orange line shows the applicability limit of exact approaches: on the left there is not enough time for the algorithm to provide a solution, the zone on the right is an area favorable for choosing exact approaches.}
        \label{fig:FirstExperiment}
    \end{figure}

    The main experimental results are presented on the Figure \ref{fig:FirstExperiment} with each algorithm represented as a point. Straight lines of different colors connect the points, showing that the overall cost of routes increases as the problem dimensions increase. The results obtained from SCIP are connected by an orange line, with blue dashes indicating the potential geometric position of the SCIP results. The x-axis represents the average solution time for a single task, while the y-axis represents the average route cost for a single task.
    
    Based on the obtained results, the following conclusions can be drawn:
    \begin{enumerate}
        \item Increasing the value of $\lambda$ in the LKH algorithm ($\lambda$ Opt) improves the quality of solutions but increases the search time.
        \item As the problem dimension and number of vehicles increase, SCIP solution time significantly increases. Doubling the problem dimension led to a roughly 14-fold increase in SCIP solution time. In contrast, the JAMPR neural network algorithm performs better than exact approaches in terms of solution quality and search time for problem instances with a dimension of 100 points. This suggests that exact methods have limited applicability for solving problems with dimensions of 100 points or more.
    \end{enumerate}
    
    The following two chapters provide a more detailed description of the conducted experiments on problem instances with dimensions of 50 and 100 points.
    
    \subsection{Results for 50-point instances}
    
    For the diagram with GAP metrics, we calculated the percentage deviation of the algorithms at each point compared to the final best result achieved within 1000 seconds of optimization (SCIP). The overall metric represents the average value among all solved problem instances.
    
\begin{figure}[!h]
    \centering
        \includegraphics[width=1.0\textwidth]{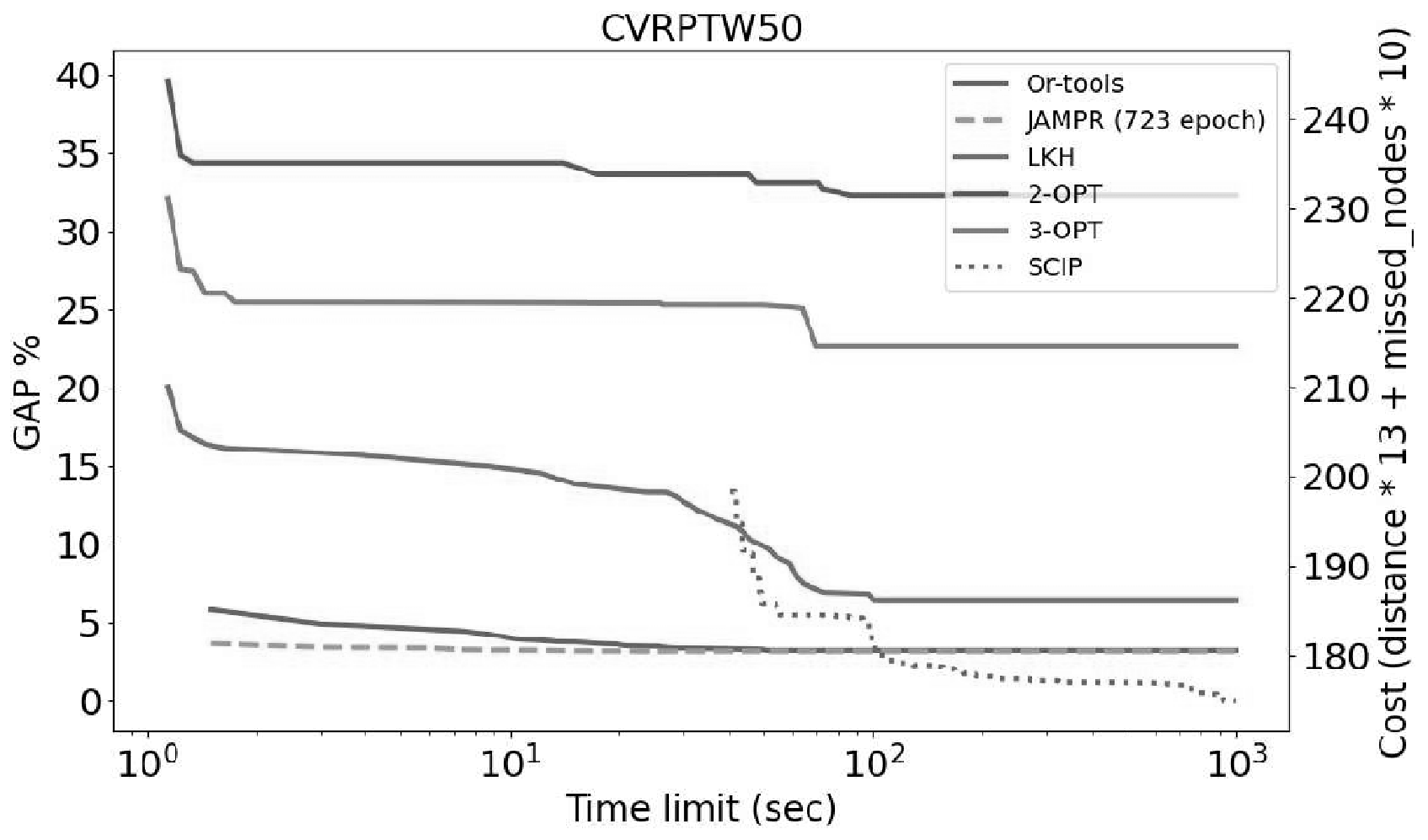}
        \caption{Performance of the considered models on the CVRPTW with 50 points. Solid lines represent the results of approximate heuristic approaches. Dotted line represents the results of SCIP - the best cost achieved with the longest time. Dashed line represents JAMPR after 700 epochs of training - a fast suboptimal solution with subsequent asymptotic approximation to the heuristic. The figure has two scales: total cost (on the right) and the gap compared to the final best solution (on the left).}
        \label{fig:cvrptw50}
    \end{figure}

    Figure \ref{fig:cvrptw50} presents the graphs showing the change in the quality of the final route costs depending on the optimization time. On the plot, it can be observed that for small 50-point instances, both classical heuristics and deep reinforcement approaches quickly provide suboptimal solutions. The results of 2-Opt, 3-Opt, LKH, OR-Tools, and JAMPR reach a plateau, but in the first few seconds of optimization, the initial greedy solutions are provided by the classical heuristics, with LKH gradually approaching the results of JAMPR and OR-Tools. However, after about 10 seconds, SCIP solution becomes inferior to LKH, OR-Tools, and JAMPR, but after 100 seconds, SCIP surpasses them. Although exact methods provide the best solution, they require significantly more time to outperform LKH, JAMPR, and OR-Tools in terms of solution quality, with the final gap between the deep reinforcement model and SCIP being less than 5\%.

    Thus, for 50-point instances, although exact approaches demonstrate the best solution, they are outperformed by classical heuristics and deep reinforcement approaches in terms of time by an order of magnitude. It is important to note that neural networks and classical heuristic approaches can provide a suboptimal solution to the CVRPTW within the first few seconds.
    
    \subsection{Results for 100-point instances}
    
    \begin{figure}[!h]
    \centering
    \includegraphics[width=1.0\textwidth]{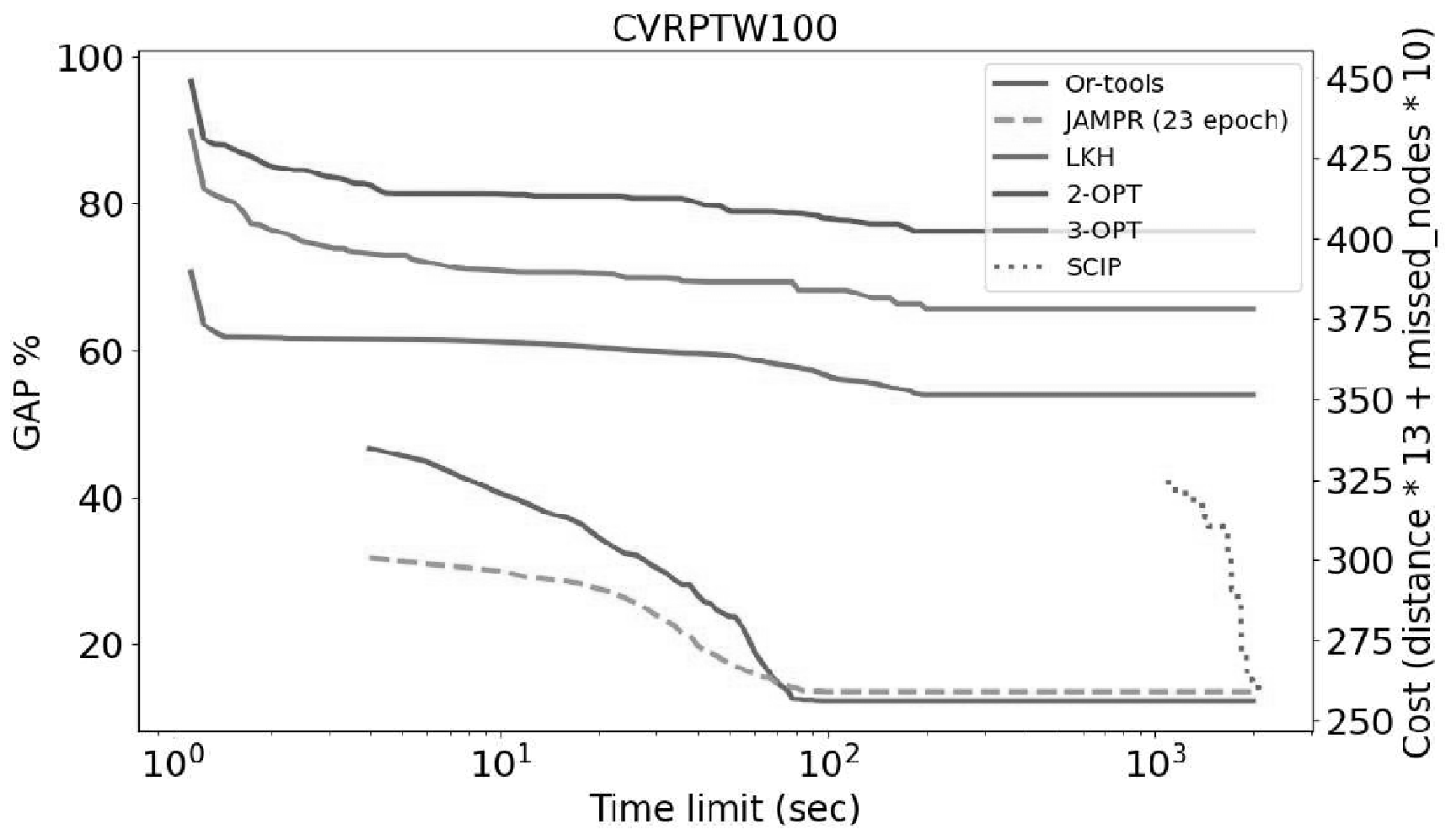}
    \caption{Performance of the considered models on the CVRPTW  with 100 points. Solid lines represent the results of approximate heuristic approaches. Dotted line represents the results of SCIP. Dashed line represents JAMPR after 23 epochs of training - a fast suboptimal solution with subsequent asymptotic approximation to the heuristic. Despite the considerable optimization time, SCIP failed to achieve the accuracy of JAMPR. The figure has two scales: total cost (on the right) and the gap compared to the final best solution (on the left).}
    \label{fig:cvrptw100}
    \end{figure}
    
    Figure \ref{fig:cvrptw100} illustrates the change in route cost quality depending on the optimization time and the chosen algorithm for 100-point instances. As with the 50-point instances, we calculated the percentage cost for each algorithm compared to the best result shown by SCIP within the maximum optimization time (in this case, 200 seconds for classical heuristics and deep reinforcement approach, and 2000 seconds for SCIP).
        
    From the graph, it is evident that classical heuristic approaches provide a faster suboptimal solution compared to other algorithms. However, they lag behind JAMPR and OR-Tools in terms of solution quality. Although JAMPR reaches a plateau around the 100th second, its initial greedy solution outperforms LKH with a GAP that is approximately 10\% better, and the final result surpasses LKH GAP by around 50\%. As the problem dimensionality doubles, there is a noticeable increasing GAP between the initial fast solutions in the deep reinforcement model and the heuristics, as depicted in Figures \ref{fig:cvrptw50} and \ref{fig:cvrptw100}.
    
    Figure \ref{fig:cvrptw100} also shows that SCIP required more than 900 seconds to obtain the first solution, which is approximately 13 times longer than it took for JAMPR and OR-Tools to achieve similar results when reaching a plateau. It is worth noting that within the allocated time, SCIP was unable to propose an optimal solution. Moreover, the first proposed solution is about 50\% more costly than the heuristic solutions provided by OR-Tools and JAMPR. This indicates that as the problem dimensionality increases, exact approaches will become less and less applicable for solving them due to exponentially increasing time.
    
    Thus, in the case of 100-point instances, deep reinforcement approaches can provide a fast suboptimal solution that outperforms classical heuristics in terms of quality due to the longer training time and outperforms exact approaches in terms of time required to find a solution.

\section{Conclusions}

    The problem of route optimization with realistic constraints is becoming extremely relevant in the face of global urban population growth. While we are aware of approaches that theoretically provide an exact optimal solution, their application becomes challenging as the problem size increases because of exponential complexity. We investigate the Capacitated Vehicle Routing Problem with Time Windows (CVRPTW) and compare solutions obtaining by exact solver SCIP \cite{BestuzhevaEtal2021OO} with heuristic algorithms such as LKH, 2-OPT, 3-OPT \cite{lkh2000}, the OR-Tools framework \cite{perron2011operations}, and the deep learning model JAMPR \cite{falkner2020learning}.

    All the metrics presented in the article were obtained using the Smart Routes platform. This system has the capability to solve the route optimization problem using various approaches, which facilitates research and result comparison.

    Based on the experimental results, the following conclusions can be drawn:
    
    \begin{itemlist}
        \item For problem instances with 50 points, both classical heuristic approaches and reinforcement learning approaches demonstrate their effectiveness by providing fast suboptimal solutions. The overall results are close to the results obtained by the exact method (SCIP) with GAP less than 5\%. This indicates that for problem instances with 50 points, classical heuristic approaches and neural network-based approaches can offer a good trade-off between solution quality and search time.
    
        \item For problem instances with 100 points, exact methods required 13 more time to find the initial solution. They were unable to provide an optimal solution within the given time, resulting in up to 50\% higher route costs during optimization. Classical heuristic approaches provide fast suboptimal solutions but start to lag behind more advanced methods such as JAMPR and OR-Tools in terms of solution quality. These findings suggest the complete infeasibility of exact methods for solving the vehicle routing optimization problem with 100 (or more) points.

    \end{itemlist}
    
    We believe that the developed platform is an important component for future research in the field of route optimization with diverse constraints. Access to the source code of the system can be obtained by communicating with the authors of the article.



\end{document}